\documentclass[final, 3p, times]{elsarticle}
\biboptions{numbers}
\usepackage[obeyspaces]{xurl} 
\usepackage{hyperref} 
\hypersetup{hidelinks,
	colorlinks=false,
	allcolors=black,
	pdfstartview=Fit,
	breaklinks=true
}

\usepackage{color}
\usepackage[nodots]{numcompress}
\usepackage{amsmath,amsfonts}
\usepackage{algorithmic}
\usepackage{algorithm}
\usepackage{verbatim}
\usepackage{float}
\usepackage{multicol}
\usepackage{multirow}

\usepackage{amssymb}

\journal{Information Sciences}
\begin{document}

\begin{frontmatter}



\title{Random Walk on Point Clouds for Feature Detection}


\author{Yuhe Zhang, Zhikun Tu, Zhi Li, Jian Gao, Bao Guo, Shunli Zhang\corref{cor1}}
\cortext[cor1]{Shunli Zhang is the corresponding author, email: slzhang@nwu.edu.cn}

\affiliation{organization={School of Information Science and Technology},
            addressline={No.1 Xuefu Road}, 
            city={Xi'an},
            postcode={710127}, 
            state={Shaanxi Province},
            country={China}}

\begin{abstract}
The points on the point clouds that can entirely outline the shape of the model are of critical importance, as they serve as the foundation for numerous point cloud processing tasks and are widely utilized in computer graphics and computer-aided design. This study introduces a novel method, RWoDSN, for extracting such feature points, incorporating considerations of sharp-to-smooth transitions, large-to-small scales, and textural-to-detailed features. We approach feature extraction as a two-stage context-dependent analysis problem. In the first stage, we propose a novel neighborhood descriptor, termed the Disk Sampling Neighborhood (DSN), which, unlike traditional spatially and geometrically invariant approaches, preserves a matrix structure while maintaining normal neighborhood relationships. In the second stage, a random walk is performed on the DSN (RWoDSN), yielding a graph-based DSN that simultaneously accounts for the spatial distribution, topological properties, and geometric characteristics of the local surface surrounding each point. This enables the effective extraction of feature points. Experimental results demonstrate that the proposed RWoDSN method achieves a recall of 0.769—22\% higher than the current state-of-the-art—alongside a precision of 0.784. Furthermore, it significantly outperforms several traditional and deep-learning techniques across eight evaluation metrics.

\end{abstract}











\end{frontmatter}

\section{Introduction}
\label{Introduction}
Point clouds, which represent 3D objects using point coordinates, are widely used in applications such as autonomous driving, robotics, computer vision, and reverse engineering\cite{ref1}. On point clouds, feature points that can completely outline the shape of the model \cite{ref2,ref3}, are crucial for point cloud processing techniques such as surface segmentation\cite{ref4}, registration\cite{PointCloudReg}, and optimization\cite{Outlier}, and can also improve deep learning tasks by improving accuracy and efficiency, as shown in \cite{StairDetection, FaceOrientationEstimation}.

Feature extraction from point clouds is traditionally viewed as a context-dependent problem, analyzing spatial \cite{ref2,ref3}, geometric \cite{ref4,ref7}, or topological \cite{ref8} information of the local neighborhood. However, existing methods struggle to encode all three types of information simultaneously into the $k$ nearest neighbors, ball neighborhood, or Delaunay neighborhood, failing to capture both large-scale textures and small-scale details. Recent deep learning approaches transformed feature detection into problems such as feature fitting \cite{ref9}, model segmentation \cite{ref10}, and parametric curve fitting \cite{ref11}, achieving state-of-the-art results. However, these methods primarily focus on coarse-grained features, such as segmentation boundaries or parametric curves, and still struggle to capture fine-scale details.

\begin{figure*}[htb]
\centering
\includegraphics[width=1.0\linewidth]{Figures/Fig0.pdf}
\caption{The results of the baseline methods and the proposed RWoDSN on three typical models with imbalanced sampling patterns (highlighted by red dotted rectangles), large-scale sharp features, small-scale smooth features (highlighted by green dotted rectangles), and small-scale detailed features (highlighted by magenta dotted rectangles), respectively.}
\label{Fig0}
\end{figure*}

In this study, we propose a method for feature detection in point clouds that captures a diverse range of characteristics, including sharp and smooth patterns as well as textural and detailed elements, as illustrated in Fig.\ref{Fig0}. We conceptualize feature detection in point clouds as a context-dependent analysis problem and address it through two key components. First, we introduce a novel neighborhood descriptor, termed the Disk Sampling Neighborhood (DSN). Subsequently, we apply a constrained random walk process to the DSN (RWoDSN), enabling feature detection by analyzing the structural properties of the resulting graph-based DSN.

In particular, the spatial distribution, geometric properties, and topological information of a neighborhood are crucial for accurately detecting feature points.
However, existing neighborhood representations, such as the $k$-nearest neighbors ($k$NN) or ball neighborhoods, are insufficient for simultaneously encoding these aspects. This limitation arises from the unstructured nature of point clouds, where kNN and ball neighborhoods tend to be scattered and lack explicit connectivity information between points. While the Delaunay neighborhood encapsulates spatial, geometric, and topological characteristics, its computational complexity is relatively high, and it often produces poor-quality triangles along the edges.

\begin{figure*}[htb]
\centering
\includegraphics[width=1.0\linewidth]{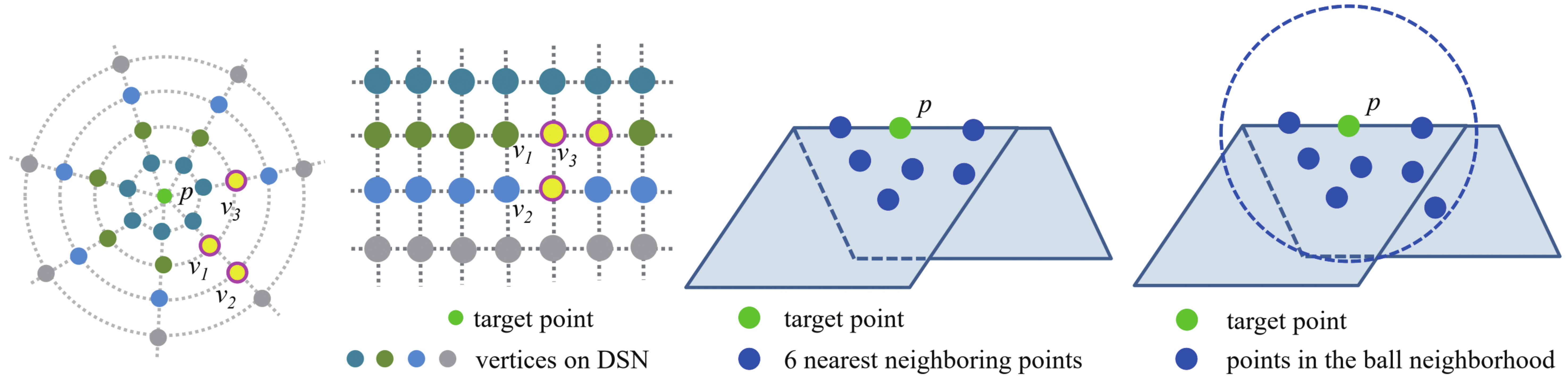}
\caption{The illustration of the proposed DSN, \textit{k}NN and the ball neighborhood.}
\label{Fig2new}
\end{figure*}

To address these challenges, we draw inspiration from the spiral aggregation map (SPLAM) \cite{ref44}, which generates a spiral aggregation map for robust template matching. Given that neighboring points closer to the target point contribute more significantly to shaping the underlying surface, we propose a radial projection method to resample points within the ball neighborhood, leading to the development of a novel neighborhood descriptor: the Disk Sampling Neighborhood (DSN). An illustration comparing DSN with kNN and ball neighborhoods is provided in Fig. \ref{Fig2new}. The DSN effectively captures all intrinsic neighborhood characteristics, including spatial distribution and shape information. Structurally, it can be represented as a matrix, where circles correspond to rows and radii to columns, preserving the topological relationships among neighboring points while maintaining structured connectivity, as shown in Fig. \ref{Fig2new}. For clarity, the term “vertex” will refer to points within the DSN, while “point” will denote points in the original point cloud in the subsequent sections.

Given the DSN, a constrained random walk process (RWoDSN) is applied to differentiate feature points from non-feature points. The core idea is based on the observation that, with a path length constraint, the set of vertices traversed by random walkers—originating from the target point—should be resampled from points lying on the same smooth underlying surface. Moreover, random walkers are less likely to traverse edges between two smooth surface patches, as these edges typically exhibit significant surface gradients. Such gradients substantially increase the path length, often exceeding the predefined constraint. Building upon this principle, by performing a restrained random walk process with a path length constraint on the DSN, the vertices visited by finite-length random walks are resampled from points on the same smooth underlying surface, ensuring connectivity within the local neighborhood. This process yields a graph-based DSN, where connectivity patterns serve as indicators of feature presence. Specifically, if the vertices in the DSN are largely or fully connected, indicating the presence of a smooth surface, the target point is classified as a non-feature point. Conversely, if only a few vertices are connected, the target point is identified as a feature point.

The primary technical contributions of this paper are two-fold:

(1) We introduce the DSN, a novel neighborhood descriptor that dynamically adapts to the local surface shape, offering a concise, structured, and compact representation. By resampling points within each bin rather than considering all neighboring points, the DSN reduces dependency on point sampling density, enhancing its robustness to varying surface geometries, noise, and imbalanced sampling patterns.

(2) We propose the RWoDSN method for feature detection on point clouds. By initiating a random walk from the target point, the process traverses vertices within the DSN, constructing a graph-based representation. Analyzing the structural properties of this graph allows for the effective identification of feature points.

(3) Experimental results demonstrate that RWoDSN effectively captures both large-scale textural features and fine-scale details, outperforming several state-of-the-art methods.

\section{Related work}
For feature extraction on 3D models, meshes \cite{ref12}, point clouds \cite{ref2,ref3}, and depth images captured from 3D models \cite{ref13}, can be used as input. We focus on feature extraction from point clouds; therefore, this section only reviews the methods that take point clouds as input.

\subsection{Non-learning-based methods}
Research into feature detection using non-learning-based methods can be divided into three main groups: those based on the analysis of point distribution, those based on the analysis of geometric invariants of point data, and those based on topological information.

Most early studies on feature detection fell into the first camp. For example, Gumhold \textit{et al}. \cite{ref2} employed covariance analysis on the coordinates of points to compute the probability that a point belongs to what kind of features; Daniels \textit{et al}. \cite{ref14} proposed a Robust Moving Least Squares (RMLS)-based feature extraction method to locally fit surfaces to separate features from non-features; Demarsin \textit{et al}. \cite{ref3} utilized the Principal Component Analysis (PCA) to deal with the point coordinates directly. 

Into the second camp falls a wide variety of methods, including curvature-based methods \cite{ref12,ref14}, normal vector-based methods \cite{ref4,ref7,ref15}, and integrated invariant-based methods \cite{ref16}. These methods usually perform diverse analysis strategies on the geometric invariants, such as tensor-voting \cite{ref17}, multi-scale analysis \cite{ref12,ref18}, clustering methods \cite{ref7}, or statistical approaches \cite{ref4,ref19}. Several other methods also combined two or more strategies to perform the analysis, for example, Weinkauf \textit{et al}. \cite{ref8} applied a topological analysis on the principal curvatures and salient edges of the surface are identified as parts of separatrices of the topological skeleton; Bazezian \textit{et al}. \cite{ref20} improved the method in \cite{ref7} by combining PCA, Gauss map clustering, and agglomerative cluster analysis for feature detection; Ahmed \textit{et al}. \cite{ref21} used an adaptive density-based threshold to detect edge points and then combined curvature vectors and their geometrical statistics to detect corner points. Liu \textit{et al.}\cite{RobustandAccurate} define a bilateral weighted centroid projection-based metric to quantify surface deviations, and then extract large to fine-scale features by iteratively performing neighborhood splitting and selection.

Several methods that address meshes reconstructed from point clouds, rather than directly working with the point clouds themselves, fall into the last camp \cite{ref8,ref25}. An alternative approach involves working with the Voronoi graph of the point cloud \cite{ref26}. However, meshes typically include underlying connectivity information associated with the vertices, while point clouds often lack such data, making feature point detection on point clouds more challenging. Moreover, the quality of mesh reconstruction and the Voronoi graph significantly influences the accuracy of feature detection results.

\subsection{Deep learning-based methods}
The rapid development of machine learning and deep learning theories makes it more efficient to detect edges on point clouds using neural networks. Early machine learning-based methods usually process the regular voxel grids of point clouds, such as \cite{ref27,ref28}. The voxelization operation entails a high computational cost and is generally limited by its resolution. Yu \textit{et al}. \cite{ref9} proposed the popular edge-aware point set consolidation Network (EC-Net) \cite{ref9}, which is designed to fit edges in point clouds rather than directly detecting edge points. Then the EC-Net\cite{ref9} was improved by Loizou \textit{et al}. \cite{ref10} to learn the boundary between different parts of a model, but the effect of sparse or tiny features is still not satisfactory. Similarly, Hu \textit{et al}. \cite{ref29} also designed the joint semantic segmentation and edge detection network (JSENet) for semantic edge detection. Wang \textit{et al.}\cite{ref11} introduced a parametric inference strategy for point cloud edges (PIE-Net), utilizing a region proposal task to detect edges as a collection of parametric curves, however, the generalization ability of parametric curves is limited as PIE-Net only deals with lines, circles, and B-spline curves. Some work improves the feature extraction performance by enriching the inputs, for example, Bazazian \textit{et al}. \cite{ref30} proposed edge detection capsule network (EDCNet) which utilizes both edge labels and segmentation labels to train the network; Himeur \textit{et al}. \cite{ref31} designed a point cloud edge detection network (PCEDNet) by adding to each point a set of differential information on its surrounding shape reconstructed at different scales. Recently, novel representations of point clouds have been introduced for feature extraction, \textit{i.e.}, Matveev \textit{et al}. \cite{ref32} proposed DEF, a learning-based framework that predicts sharp geometric features in sampled 3D shapes by regressing a scalar field representing the distance from point samples to the closest feature line on local patches; Zhu \textit{et al}. \cite{ref33} presented NerVE\cite{ref33}, which consists of a novel neural volumetric edge representation and a volumetric learning framework; Ye \textit{et al}. \cite{ref34} presented NEF learning a neural implicit field based on Neural Radiant Field (NeRF). 

\subsection{Random walk}
Random Walk is a mathematical formalization of a path consisting of a series of random steps \cite{ref35}, which has been used for image segmentation \cite{ref36}, point cloud segmentation \cite{ref37}, and community detection \cite{ref38}. Recently, there has been an active interest in representing point clouds or meshes using random walks for combining with deep learning. For example, proposed by Lahav \textit{et al}., MeshWalker\cite{ref39} first uses random walk to obtain a representation of point clouds and then feeds this representation into a Recurrent Neural Network (RNN) framework for mesh classification and mesh semantic segmentation. Similarly, random walk has also been utilized to encode point clouds into patches \cite{ref40} or curves \cite{ref41}, which are taken as the input of the network or a module in the network, for mesh shape analysis. Besides object classification, retrieval, and segmentation, random walk has also been utilized to refine the pseudo labels for scene flow estimation from point clouds \cite{ref42}.

\subsection{Summary}

The key differences between the proposed method and existing feature extraction techniques are as follows. Traditional non-learning-based methods mainly focus on spatial distribution, geometric invariants, or connectivity on reconstructed meshes, often failing to integrate these aspects comprehensively. This limits their ability to handle features at varying scales. In contrast, our approach incorporates spatial distribution alongside the combination of connectivity and geometry using a graph-based neighborhood descriptor, eliminating the need for mesh reconstruction. This enables more effective extraction of diverse features across different scales. Additionally, existing deep learning-based methods often frame feature detection as surface segmentation, boundary detection, or parametric curve detection, which typically results in lower accuracy when capturing fine-scale features.

Unlike existing random-walk methods for point clouds, our approach does not require meshes, voxels, or surface reconstruction, even though the DSN retains a matrix structure. Existing random walk-based methods primarily focuse on large-scale features, such as shape classification, retrieval, and semantic segmentation. In contrast, feature extraction from point clouds is typically concerned with fine-scale information, where individual points are the focal points. Small-scale noise and imbalanced local sampling densities can significantly impact individual points, whereas large-scale noise primarily affects the overall shape. As a result, feature extraction from point clouds is more susceptible to noise and local sampling density, making it a more challenging task. 


\section{Method}
In this study, we developed a novel neighborhood descriptor and a simple feature detection algorithm for detecting features in point clouds. We first introduced an efficient point resampling scheme to form the neighborhood descriptor, DSN. Since point clouds consist of scattered points without underlying connectivity information, we divided the spatial region around each point into several bins and resampled the points within each bin to create the descriptor. This approach enables the DSN to capture both the spatial distribution and the overall geometry of the neighborhood around each point, while retaining a grid-like structure that can be represented as a matrix. As a result, the DSN simultaneously encodes spatial distribution, geometric, and topological information.

The primary goal of this study was to develop an efficient feature detection method. For each point’s DSN, we performed a constrained random walk process. The vertices traversed by the walker were then connected to their neighbors, resulting in a graph-based DSN. This graph is subsequently analyzed to identify the feature points.

\subsection{The DSN construction}
The construction of the DSN involves resampling the points within the ball neighborhood around a point on the point cloud. Unlike directly dividing or connecting the neighboring points on the point cloud, as seen in \cite{ref43}, the proposed DSN acts as a mapping from the original point cloud to the disks. These disks consist of a set of centroids of points that fall within the corresponding angle and Euclidean distance.

\begin{figure*}[htb]
\centering
\includegraphics[width=1.0\linewidth]{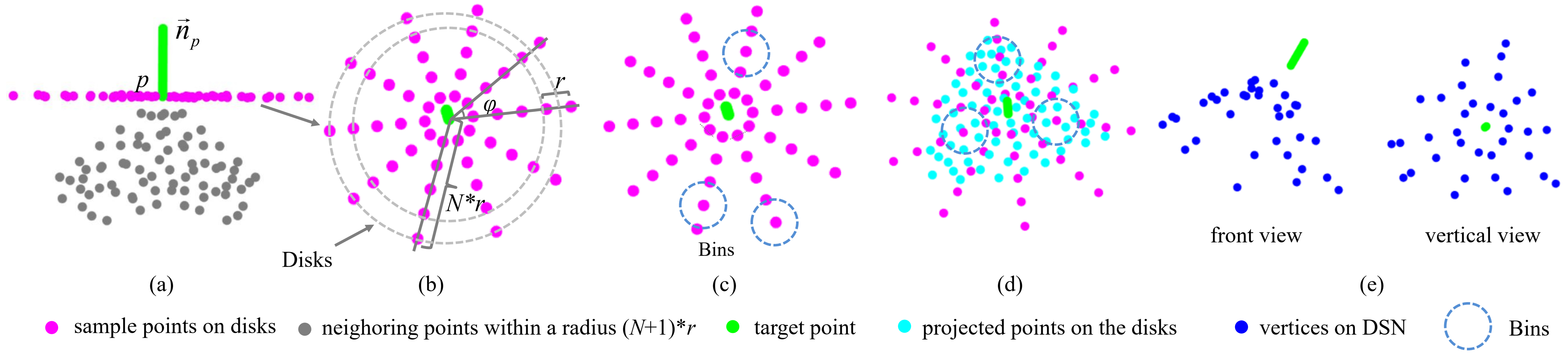}%
\caption{The illustration of the DSN construction for point $p$. (a) A series of nested disks centered at point $p$ and perpendicular to the normal vector. (b) The parameters that determine the sampling points on the disks. (c) The bins on the disk plane. (d) The neighboring points of $p$ projected onto the disk plane. (e) The front view and side view of the DSN for point $p$.}
\label{Fig1}
\end{figure*}

Given a point cloud $P=\{p_i\}\in R,0 \leq i \leq N$, let $p$ denote a point on the point cloud, and $\vec{n}_p$ (green line in Fig. \ref{Fig1}) represents the normal vector of point $p$. The normal vector of a point can be estimated using various techniques\cite{ref45,ref46}. In this work, the normal vector of the point is estimated using Meshlab, as the estimation method is not the focus of this study.

The construction of the DSN of $p$, illustrated in Fig. \ref{Fig1}, can be divided into two steps. First, a series of concentric circles are created, forming the nesting disks. Specifically, the $N$ disks are all centered at the target point $p$ and are perpendicular to the normal vector $\vec{n}_p$, as shown in Fig. \ref{Fig1}(a)-(b). The radii difference between the disks is set to a default parameter $r=S_{density}$. The radius of the innermost circle is $r$, and the difference between the radii of adjacent circles is also $r$, so the radius of the second circle is $2 * r$, and so on. $S_{density}$ is computed using Equ.\eqref{S_density}:

\begin{equation}
\label{S_density}
S_{density}=\frac{1}{N_p}\sum_{p_j\in Np}\left\| p-p_j\right\|_2
\end{equation}

\noindent where $\| \bullet \|_2$ is the Euclidean distance and $N_p$ is the $k$ nearest neighbors.

The angle $\varphi$ between two adjacent sampling points on the disks (magenta points in Fig. \ref{Fig1}) determines the sampling granularity on each disk. The value of $\varphi$ is consistent across all disks, leading to ${2\pi}/{\varphi}$ sets of points on each disk, as shown in Fig.\ref{Fig1}(b). Thus, both the number of the disks $N$ and the angle $\varphi$ govern the scale of the DSN, resulting in $N*{2\pi}/{\varphi}$ bins, as shown in Fig.\ref{Fig1}(c). 

\begin{figure}[htb]
\centering
\includegraphics[width=\linewidth]{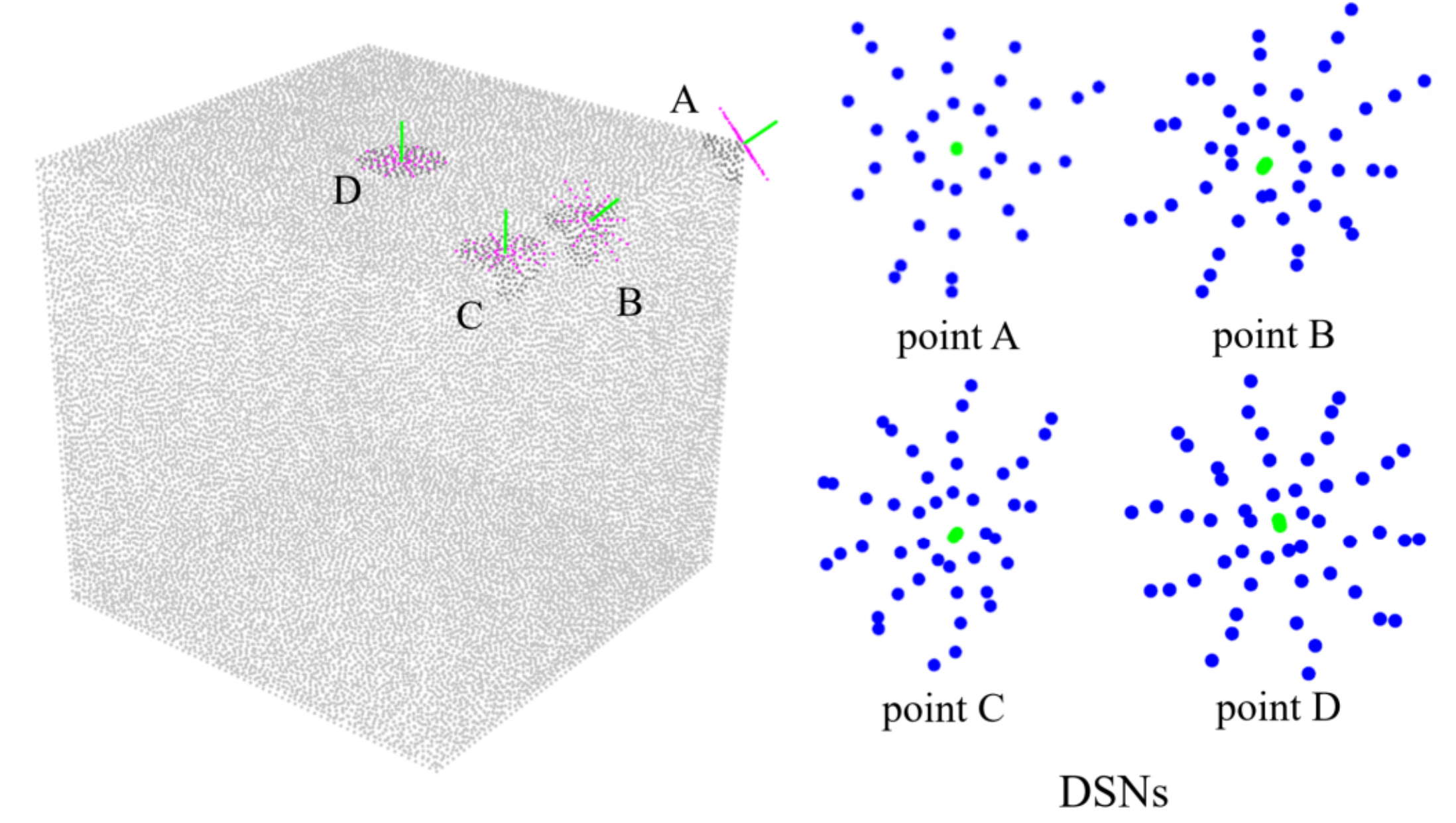}
\caption{The DSN of four different points: point A is a corner point, point B is an edge point, point C is a non-feature point near the edge, and point D is a non-feature point on a smooth surface.}
\label{Fig4new}
\end{figure}

Second, given the disks, all neighboring points $p_j$ within a radius of $(N+1) * r $ (gray points in Fig. \ref{Fig1}(a)) are projected onto the disk plane, where $N$ is the number of disks. We use the ball neighborhood rather than \textit{k}NN, as \textit{k}NN may be biased towards one side of the target point due to imbalanced sampling in point clouds, as shown in Fig.\ref{Fig2new}. Let $p_j'$ denote the projected point (cyan points in Fig. \ref{Fig1}(d)) of $p_j$. The point $p_j$ is then classified into the corresponding bin, depending on where $p_j'$ falls. More precisely, if the projected point $p_j'$ satisfies the constraint in Equ.\eqref{bins}, then the point $p_j$ is assigned to the bin centered at $c_t$ ($0 \leq t \leq {2\pi}/{\varphi}-1$), where $c_t$ is the sample point on the disks (magenta points in Fig. \ref{Fig1}(c)-(d)). 

\begin{equation}
\label{bins}
 \left\|p_j'-c_t \right\|_2 \leq r
\end{equation}   

\noindent where $\| \bullet \|_2$ is the Euclidean distance. After all neighboring points are classified, the centroid of the points within each bin is computed and considered a vertex on the DSN (blue points in Fig. \ref{Fig1}(e)). Finally, all $N*{2\pi}/{\varphi}$ centroids (instead of the individual neighboring points $p_j$) form the DSN of the target point $p$. The DSN of four different points is illustrated in Fig.\ref{Fig4new}.

\subsection{RWoDSN for feature detection}

A point can be classified as a “feature point” or “non-feature point”, by analyzing its neighborhood. Similarly, the analysis can be extended to the DSNs to detect feature points. Therefore, we propose the Random Walk on DSN (RWoDSN) feature extraction approach, which applies the random walk process to the DSNs.

\begin{figure*}[htb]
\centering
\includegraphics[width=1.0\linewidth]{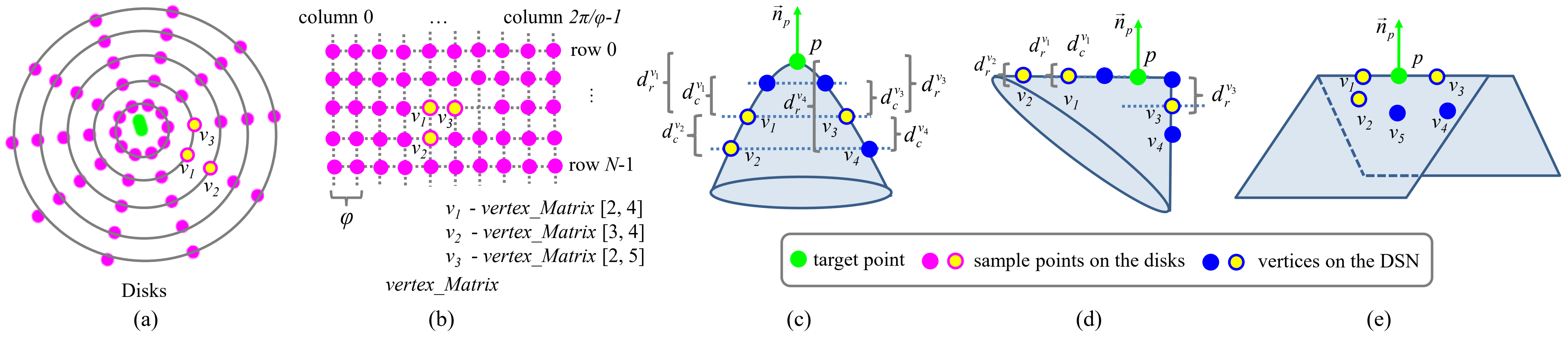}%
\caption{The illustration of $vertex\_Matrix$, $d_r^{i,j}$ and $d_c^{i,j}$: (a) shows the nesting disks, which can be considered as an $N\times 2 \pi / \varphi$ matrix, as shown in (b). (c) shows an example of a non-feature point on a smooth surface, (d) shows an example of a non-feature point located near the edge, and (e) shows an example of an edge point.}
\label{Fig4}
\end{figure*}

The DSN constructed in the previous step could be regarded as a matrix structure, as shown in Fig.\ref{Fig4}. Thus, the vertices on the DSN can be uniquely positioned in the matrix with a fixed size of $N\times 2\pi/\varphi$. Consequently, the vertices on the DSN are stored in a matrix referred to as $vertex\_Matrix$. More specifically, the vertices on the same disk(cf. Fig.\ref{Fig4}(a)-(b)) are considered to be in the same row of the $vertex\_Matrix$, while vertices on the same radius are placed in the same column, thereby retaining a regular neighboring structure. For example, a vertex at $vertex\_Matrix[i,j]$ has four direct neighbors, namely $vertex\_Matrix[i,j-1]$, $vertex\_Matrix[i,j+1]$, $vertex\_Matrix[i+1,j]$ and $vertex\_Matrix[i-1,j]$. Note that vertices in column 0 and column $2\pi/\varphi-1$ are also the direct neighbors, and if $i=0$, the $vertex\_Matrix[i-1,j]$ neighbor corresponds to the target point. 

The vertices on the DSN of a non-feature point $p$ that lies on a smooth surface are resampled from points on the same underlying smooth surface; an example is illustrated in Fig.\ref{Fig4}(c). Consequently, the vertices in the same row should exhibit similar distances $d_r^v=d_r^{i,j}=\left\| v-{v}' \right\|_2$, where ${v}'$ is the projected point of $v$ on the disk plane, and $i$ and $j$ are the indices of the vertex $v$ in the $vertex\_Matrix$. Moreover, vertices located in the same column and adjacent rows (\textit{e.g.} $vertex\_Matrix[i-1, j]$ and $vertex\_Matrix[i, j]$, with $1 \leq i\leq N-1$) also have nearly equal intervals $d_c^v$, which are defined as follows:

\begin{equation}
\label{dc}
d_c^v=d_c^{i,j}=\left\{\begin{array}{l}
|d_r^{i,j}-d_r^{i-1,j}|, \quad i\neq 0 \\
|d_r^{i,j}-d_r^{target\_point}|, \quad i= 0 
\end{array}\right.
\end{equation}      

\noindent where $i$ and $j$ represent the indices of the vertex $v$ in the $vertex\_Matrix$. For example, for the non-feature point $p$ in Fig.\ref{Fig4}(c), the DSN of $p$ has $d_r^{v_1}\approx d_r^{v_3}$, $d_r^{v_2}\approx d_r^{v_4}$, $d_c^{v_1}\approx d_c^{v_2}$ and $d_c^{v_3}\approx d_c^{v_4}$. However, as shown in Fig.\ref{Fig4}(d), for a non-feature point $p$ located near the edge, some of the vertices on the DSN of $p$ are resampled from points that lie on the same underlying smooth surface as the target point $p$, such as vertices $v_1$ and $v_2$, while other vertices are resampled from points on adjacent surfaces, for example, vertices $v_3$ and $v_4$. Therefore, based on the previously defined $d_r^{i,j}$ and $d_c^{i,j}$, for the DSN of the non-feature point $p$ in Fig.\ref{Fig4}(d), located near the edge, $d_r^{v_1}\neq d_r^{v_3}$, $d_r^{v_2}\neq d_r^{v_4}$ and $d_c^{v_1}\approx d_c^{v_2}$. Similarly, for the vertices on the DSN of a feature point $p$, only a few vertices resampled from the points in the edge region, may have similar $d_r^{i,j}$, \textit{e.g.}  $d_r^{v_1}\approx d_r^{v_3}$, whereas $d_r^{i,j}$ and $d_c^{i,j}$ of the other vertices are random values. An example is illustrated in Fig.\ref{Fig4}(e). 

Therefore, two vectors are built for storing the differences between $d_r^{i,j}$ and $d_c^{i,j}$ of direct neighboring vertices. These vectors are used for analyzing the location of the target point and further identifying features. The first vector, denoted as $rowVector$ in Equ.\eqref{rowVector}, stores the $d_r^{i,j}$ differences between a vertex in $vertex\_Matrix(i,j)$ and its directly adjacent neighbor in $vertex\_Matrix(i, j-1)$. The second vector, $colVector$ in Equ.\eqref{columnVector}, stores the $d_c^{i,j}$ differences between vertices in the same column and adjacent rows:

\begin{equation}
\label{rowVector}
\begin{split}
rowVector(v)=rowVector\left [ (2\pi/\varphi -1)*i+j \right ]\\
 =\left\{\begin{array}{l}
|d_r^{i,j}-d_r^{i,j-1}|,\quad j \neq 0 \\
|d_r^{i,j}-d_r^{i,2\pi/\varphi -1}|,\quad j = 0
\end{array}\right. 
\end{split}
\end{equation}

\begin{equation}
\label{columnVector}
\begin{split}
colVector(v)=colVector\left [ (2\pi/\varphi -1)*i+j \right ]\\
=\left\{\begin{array}{l}
|d_c^{i,j}-d_c^{i-1,j}|,i \neq 0 \\
|d_r^{i,j}-d_r^{target\_point}|,i = 0
\end{array}\right. 
\end{split}
\end{equation} 

\noindent where $i$ and $j$ are the indices of vertex $v$ in $vertex\_Matrix$, and $rowVector$, and $colVector$ store the differences between $d_r^{i,j}$ and $d_c^{i,j}$ of direct neighboring vertices. These vectors can also be interpreted as representing the edge weights between the vertices and their direct neighbors.

The values in $rowVector$ and $colVector$ indicate the edge weights between vertices, which can be used to compute the path length of a random walk on the DSN. By setting a path length constraint, the random walk process, which starts from the target point $p$, tends to favor vertices resampled from points on the same underlying smooth surface. This is because the path length through vertices resampled from the same smooth surface is typically short, while the path length increases significantly when the walker reaches vertices resampled from points near the edge. Consequently, the vertices traversed by the random walker are connected, forming a graph-based DSN. By analyzing the graph-based DSN, the location of the target point can be effectively identified, and feature points can be distinguished from non-feature points based on their locations.

\begin{algorithm}[H]
\caption{RWonDSN}
\label{RWoDSN Algorithm}
\begin{algorithmic}
\STATE 
\STATE {01: Input: DSN of point $p$}
\STATE {02: for $s=0$ to $T-1$}
\STATE {03:\quad start from the target point $p$}
\STATE {04:\quad select vertex $v_a$ with minimum ${colVector}[v_a]$\\ \quad \qquad and $v_a$ is in $0th$ row in $vertex\_Matrix$ }
\STATE {05:\quad $d_{mean}^{path} = colVector[v_a]$}
\STATE {06:\quad $Num=1$}
\STATE {07:\quad $Sum_{path} = colVector[v_a]$}
\STATE {08:\quad for $step=1$ to m}
\STATE {09:\qquad if no neighboring vertex $v_b$ meets Equ.\eqref{pathConstraint}}
\STATE {10:\qquad break}
\STATE {11:\qquad randomly select a neighboring vertex $v_b$ of $v_a$}
\STATE {12:\qquad if $v_b$ meets Equ.\eqref{pathConstraint}}
\STATE {13:\qquad \quad $step++$}
\STATE {14:\qquad \quad $Counting\_matrix(v_a,v_b)++$}
\STATE {15:\qquad \quad $v_a=v_b$}
\STATE {16:\qquad \quad $Num++$}
\STATE {17:\qquad \quad $Sum_{path}+=vector[v_b]$}
\STATE {18:\qquad \quad update $d_{mean}^{path} = Sum_{path}/{Num}$}
\STATE {19:\qquad end if}
\STATE {20:\quad end for}
\STATE {21: end for}
\STATE {22: Outputs: $Counting\_matrix$}
\end{algorithmic}
\label{alg1}
\end{algorithm}

Hence, a constrained random walk process with a path length constraint, starting from the target point $p$, is performed on the DSN to detect the feature points. The detailed procedure of the random walk is outlined in Algorithm \ref{RWoDSN Algorithm}. In steps 8–20 of Algorithm \ref{RWoDSN Algorithm}, $m$-step random walks  are executed from the target point $p$ on the DSN, and the $Counting\_matrix$, tracks the number of times each point is visited by the random walkers, is generated. The path length constraint is defined as in Equ.\eqref{pathConstraint}, the $S_{density}$ value is computed using Equ.\eqref{S_density}, and $d_{mean}^{path}$ represents the average path length of the current random walk.

\begin{equation}
\label{pathConstraint}
\left\| vector[v_b] - d_{mean}^{path} \right\|_2 \leq 0.1 * S_{density}
\end{equation}

\noindent where $vector(v_b)$ refers to either $rowVector(v_b)$ or $colVector(v_b)$, depending on the adjacent relationship between $v_a$ and $v_b$. 

\begin{figure}[htb]
\centering
\includegraphics[width=0.7\linewidth]{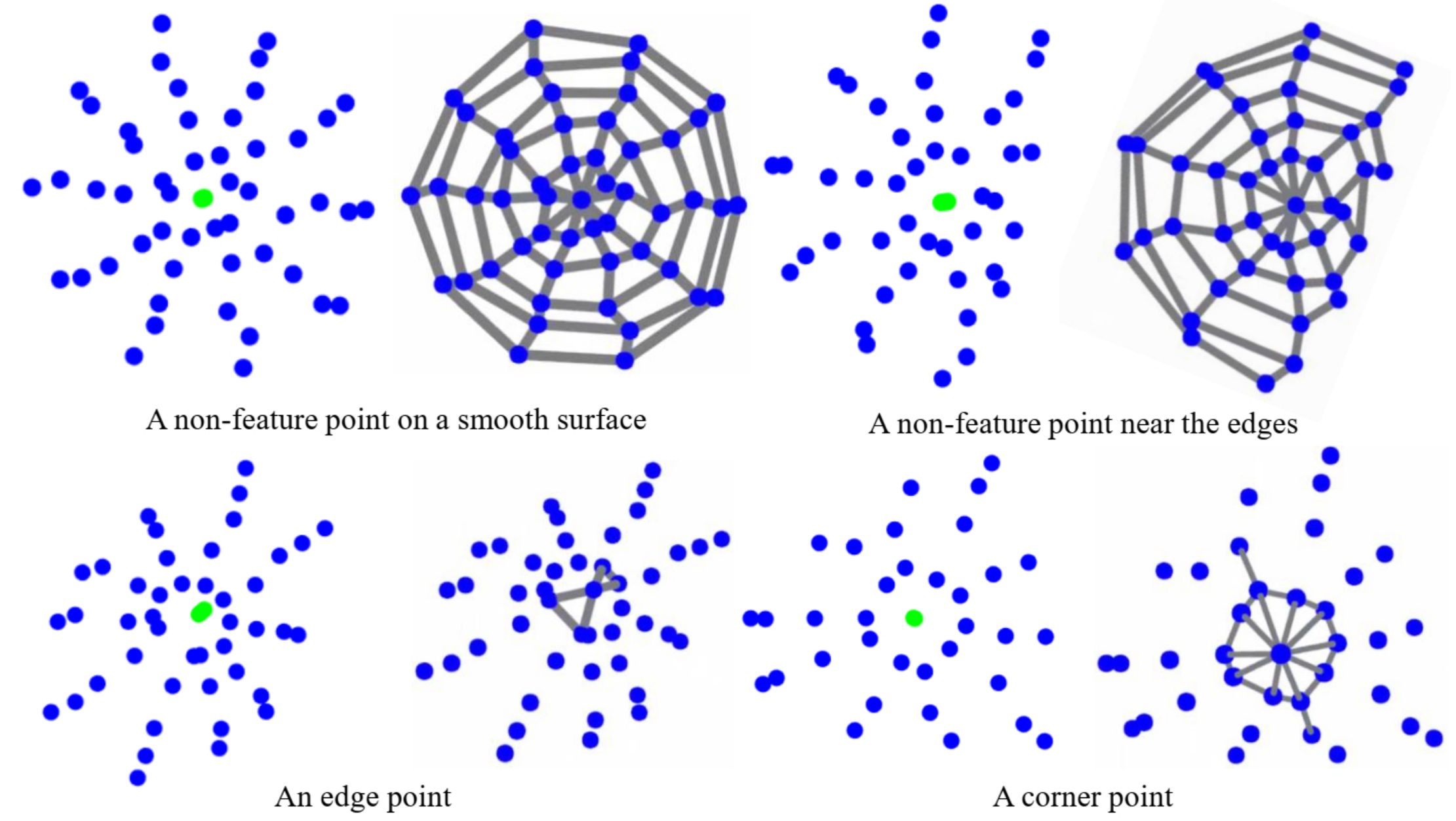}
\caption{The resulting graph-based DSNs of the points in Fig.\ref{Fig2new}.}
\label{Fig5}
\end{figure}

Ideally, by applying Algorithm \ref{RWoDSN Algorithm}, the resulting graph would connect all the vertices that the random walker has passed. However, due to the potential for random walkers to move around only a subset of vertices, some vertices may be omitted. To address this, random walks starting from the target point $p$ are repeated $T$ times (steps 2–21). Only the vertices that are frequently visited by the walkers are connected. In the experiments presented in this paper, a connection is established between vertices $v_a$ and $v_b$ in the resulting graph-based DSN if $Counting\_matrix(v_a,v_b) \geq T/3$. The graph-based DSNs for four different points are shown in Fig.\ref{Fig5}.

As shown in Fig.\ref{Fig5}, by applying Algorithm \ref{RWoDSN Algorithm}, the vertices on the DSNs of non-feature points located on smooth surfaces are fully connected. For non-feature points near the edges, the vertices on the DSNs are partially connected, while only a few vertices on the DSNs of edge and corner points are connected.

In the graph-based DSN generated by RWoDSN, a vertex connected to at least one other vertex is considered a non-isolated vertex, while a vertex with no edges connecting it to its adjacent vertices is considered an isolated vertex. Based on the graph-based DSN generated by RWoDSN, non-feature points can be identified under the following two constraints, while the remaining points are classified as feature points.

\noindent \textbf{Constraint 1: Non-feature points on the smooth surface.} If, in the DSN, the number of the non-isolated vertices is greater than $0.9 * N * 2\pi / \varphi -1$, the point is considered a non-feature point on the smooth surface.

\noindent \textbf{Constraint 2: Non-feature points near the edges.} If the graph-based DSN generated by RWoDSN satisfies either  Equ.\eqref{rowConn} or Equ.\eqref{columnConn}, the vertex is considered  a non-feature point near the edges.

\begin{equation}
\sum_{i=0}^{N-1}Con_i \geq N-2
\end{equation}

\begin{equation}
\label{rowConn}
Con_i=\left\{
\begin{array}{l}
1,\quad rowCon\_max_i >0.5*2\pi/\varphi-2\\
0,\quad otherwise
\end{array}\right.
\end{equation}

\begin{equation}
\sum_{j=0}^{2\pi /\varphi -1}Con_j \geq 0.5*2\pi/\varphi-2
\end{equation}

\begin{equation}
\label{columnConn}
Con_j=\left\{
\begin{array}{l}
1, \quad columnCon\_max_j >N-2 \\
0, \quad otherwise
\end{array}\right.
\end{equation}

\noindent where $rowCon\_{max_i}$ represents the maximum number of  continuously connected vertices in the $ith$ row of the $vertex\_Matrix$, and  $columnCon\_{max_j}$ denotes the maximum number of continuously connected vertices in the $jth$ column of the $vertex\_Matrix$.

\section{EXPERIMENTAL RESULTS}
\subsection{Experimental Settings}

\subsubsection{Datasets} 

The ABC dataset \cite{ref47} is the largest publicly available dataset of CAD models and has been widely used to test the performance of feature extraction techniques in various studies \cite{ref9,ref11,ref32,ref33,ref34}. The ground truth for the models is based on their decomposition into individual patches, making it well-suited for evaluating the performance of edge detectors and part boundary detectors. Furthermore, the ABC dataset contains models with both sharp and smooth features, and all the models exhibit non-uniform sampling densities. Comprising a total of one million manually-created CAD models, we randomly selected 6000 models with the prefix "0000" for our experiments. These models feature different types of geometries, including lines, circles, ellipses, B-splines, and others. Moreover, some widely used real-scanned models, such as the block model, the trim-star model, and the renowned Fandisk model, were also incorporated for testing.

\subsubsection{Baseline methods} 

We compare our RWoDSN with several traditional methods, including BE (Boundary Estimation) implemented in Point Cloud Library (PCL)\cite{ref48}, PBRG\cite{ref4}, and SGLBP\cite{ref25}, as well as with three typical deep learning networks: EC-Net\cite{ref9}, PIENet\cite{ref11} and NerVE\cite{ref33} which serve as baselines. BE is a widely-used feature estimation method in PCL \cite{ref48}, while PBRG \cite{ref4} applies Poisson distribution analysis on point normals for feature detection. Similar to PBRG, our method also utilizes normal vector information. SGLBP \cite{ref25} first reconstructs local meshes from point clouds and then performs neighborhood analysis on sub-graphs derived from these meshes. EC-Net\cite{ref9} was introduced for edge detection on point clouds and 3D reconstruction. It is trained on the shapeNet dataset and has shown 
satisfactory results and generalization ability on the ABC dataset. PIENet\cite{ref11}, or Parametric Inference of Edge, is trained specifically on the ABC dataset. NerVE\cite{ref33} first voxelized the point clouds and then learns features based on the neural volumetric edge representation. 

For testing, we use the provided codes for PBRG\cite{ref4}, SGLBP\cite{ref25}, EC-Net\cite{ref9}, PIENet\cite{ref11} and NerVE\cite{ref33}, as well as the BE implemented in PCL\cite{ref48}, to detect features. Our proposed RWoDSN is implemented in C++ using PCL\cite{ref48}. In our experiments, the point clouds are down-sampled to 8096 points through uniform sampling (implemented in open3D) for PIENet\cite{ref11}, as described in \cite{ref11}, while the original point clouds without any preprocessing are used as input for the other baseline methods and our method.

\subsubsection{Parameters} 

For the BE, PBRG\cite{ref4}, SGLBP\cite{ref25}, EC-Net\cite{ref9}, PIENet\cite{ref11} and NerVE\cite{ref33}, we use the default parameters suggested in their respective references. The parameters that determine the scale and sampling granularity of the proposed DSN are $\varphi$ and $N$. In our experiments (except for the experiments in Section \textit{Sensitiveness to parameters}), we set $\varphi=36^{\circ }$, which results in 10 sampling points on each disk, and $N=5$, which indicates 5 disks. Consequently, the number of vertices on the DSN is 50. 

\subsection{Evaluation metrics}
Inspired by the literature on feature curve extraction from triangle meshes\cite{ref49} and SGLBP\cite{ref25}, we employ nine metrics to evaluate the performance of the selected methods. Since EC-Net\cite{ref9} and PIENet\cite{ref11} fit features rather than directly extracting the original feature points from point clouds, we normalize the outputs (feature point cloud $P_f$) of all methods along with the Ground Truth ($GT$). Additionally, the nearest points are identified as corresponding point pairs calculated using Iterative Closest Points (ICP) \cite{ref25}, ensuring improved accuracy in the evaluation results. The metrics used are listed as follows:

\textbf{Hausdorff distance.} The Hausdorff distance between the point sets $P_f$ and $GT$ is computed as follows\cite{ref49}:

\begin{equation}
d_{Hausd}(P_f,GT)=max_{(a \in P_f)}min_{(b \in GT)}d(a,b)
\end{equation}

\noindent with $d$ is the Euclidean distance, $a$ and $b$ are corresponding point pairs obtained by ICP. The Hausdorff distance between $P_f$ and $GT$ is:

\begin{equation}
max\{d_{Hausd}(P_f,GT),d_{Hausd}(GT,P_f)\}
\end{equation}

\textbf{\textit{Jaccard.}} The second metric is \textit{Jaccard} as defined in Equ.\eqref{Jaccard} \cite{ref49}, which is utilized to evaluate how well the features extracted by the participants fit the Ground Truth and vice-versa:

\begin{equation}
dice(P_f,GT)=\frac{2\left|P_f \cap GT \right|}{\left| P_f\right|+\left| GT\right|}
\end{equation}

\begin{equation}
Jaccard(P_f,GT)=\frac{dice(P_f,GT)}{2-dice(P_f,GT)}
\label{Jaccard}
\end{equation}

\noindent $|P_f|$  represents the points’ number in $P_f$, and $|GT|$ represents the points’ number in Ground Truth. $|P_f \cap GT|$ represents the corresponding point pairs’ number between $P_f$ and GT.

\textbf{Matthews Correlation Coefficient (\textit{MCC}).} \textit{MCC} is a correlation between the observed and predicted binary classification and ranges from $[-1,1]$, which is calculated based on true positives(\textit{TP}), false positives(\textit{FP}), true negatives(\textit{TN}) and false negatives(\textit{FN}), and we regard the corresponding point pairs achieved by ICP as \textit{TP}. 

\begin{equation}
MCC=\frac{TP \times TN - FP \times FN}{\sqrt{(TP+FP)(TP+FN)(TN+FP)(TN+FN)}}
\label{MCC}
\end{equation}

\textbf{Intersection over Union Score (\textit{IoU}).} \textit{IoU} specifies the overlap between the $P_f$ and GT; which is also calculated using \textit{TP}, \textit{FP}, \textit{TN} and \textit{FN}. 

\begin{equation}
IoU=\frac{TP}{TP+FP+FN}
\label{IoU}
\end{equation}

\textbf{Precision.} Precision which is calculated using Equ.\eqref{precision}, which measures the precision of the methods. 

\begin{equation}
Precision=\frac{|P_f \cap GT|}{|P_f|}
\label{precision}
\end{equation}

\textbf{Recall.} Recall is calculated using Equ.\eqref{recall}, which measures the recall of the methods.

\begin{equation}
Recall=\frac{|P_f \cap GT|}{|GT|}
\label{recall}
\end{equation}

\textbf{\textit{MSE} (Mean Squared Error)}. \textit{MSE} is the mean squared error of the distance between the corresponding point pairs, which is calculated using Equ.\eqref{MSE}.

\begin{equation}
MSE=\frac{\sum(a-b)^2}{|P_f \cap GT|},(a \in P_f, b \in GT)
\label{MSE}
\end{equation}

\textbf{$d_{max}$}. $d_{max}$ is the maximum Euclidean distance between the corresponding point pairs which is calculated using Equ.\eqref{dmax}. 

 \begin{equation}
d_{max}=max(||a-b||_2), (a \in P_f, b \in GT)
\label{dmax}
\end{equation}

\textbf{$d_{min}$}. $d_{min}$ is the maximum Euclidean distance between the corresponding point pairs which is calculated using Equ.\eqref{dmin}. 

 \begin{equation}
d_{min}=min(||a-b||_2), (a \in P_f, b \in GT)
\label{dmin}
\end{equation}

Amongst these evaluation metrics, the lower the Hausdorff distance, \textit{MSE}, $d_{max}$, and $d_{min}$ are (0 at best), the better. However, for the \textit{Jaccard}, \textit{MCC}, \textit{IoU}, precision, and recall, the higher the scores are (1 at best), the better. 

\subsection{Sensitivity to parameters}

The main parameters used in RWoDSN are $\varphi$ and $N$, both of which determine the scale and sampling granularity of the DSN. A larger value of $\varphi$ leads to fewer vertices in the DSN, while a larger value of $N$ results in more vertices. In this section, we evaluate the sensitivity of the proposed RWoDSN to parameters by setting different values for $\varphi$ and $N$ during feature extraction on various models.

\begin{figure}[H]
\centering
\includegraphics[width=1.0\linewidth]{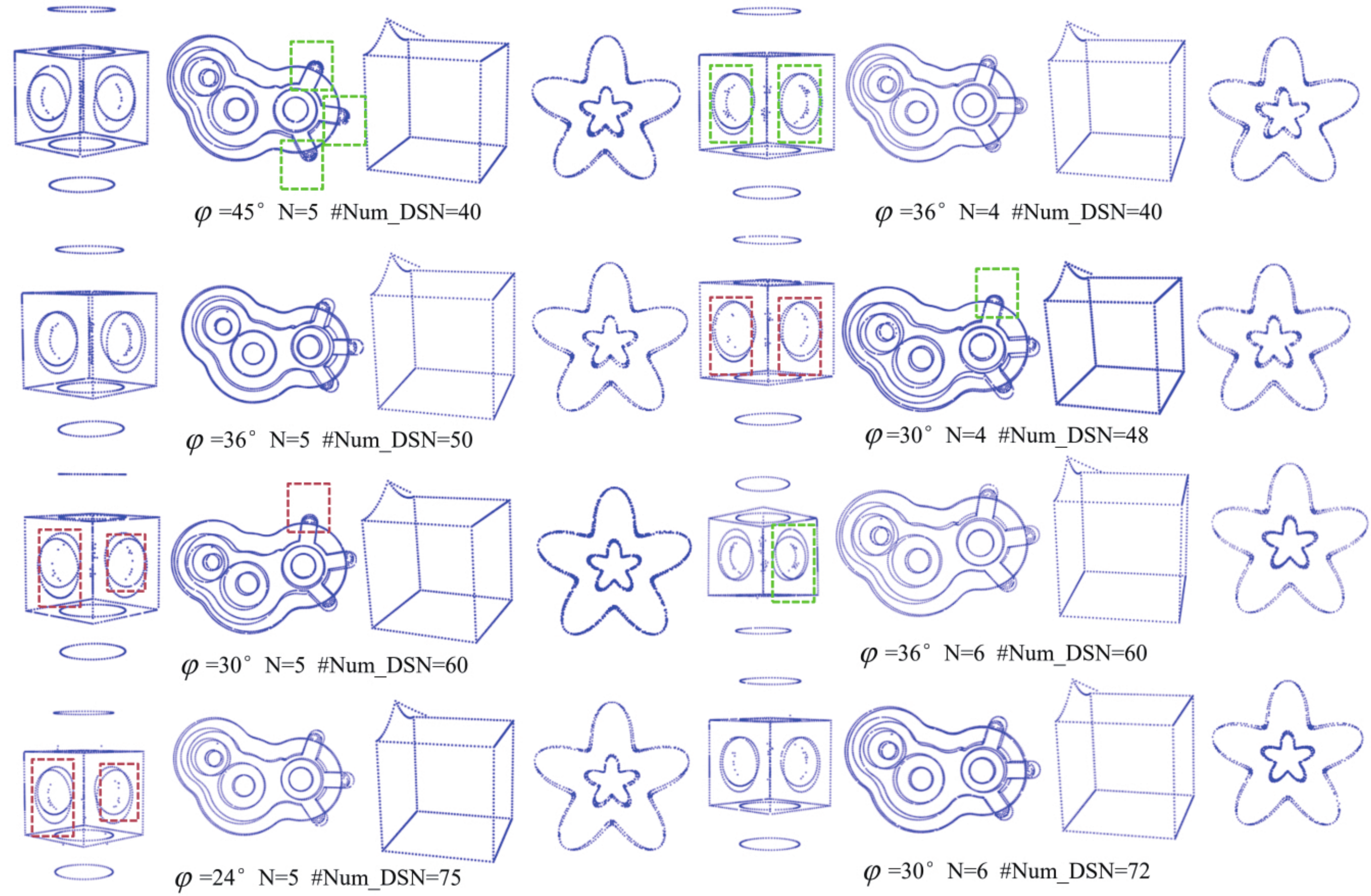}
\caption{The results of our RWoDSN using different parameters (with models and ground truth are shown in Fig.\ref{Fig9}).}
\label{Fig7}
\end{figure}

Fig.\ref{Fig7} illustrates the results of the proposed RWoDSN using different parameter values, with the scale of the resulting DSN ranging from 40 to 75. The four tested models feature various types of elements, including both sharp and smooth features. First, we fix the value of
$N$ (\textit{i.e.} $N=5$) and adjust $\varphi$ with the results displayed in the left four columns of Fig.\ref{Fig7}. Next, we performed two sets of experiments using different combinations of $\varphi$ and $N$ values, such as $\varphi=36^{\circ}$ or $\varphi=30^{\circ}$ with $N=4$ or $N=6$, with the corresponding results shown in the right four columns of Fig.\ref{Fig7}. 

As observed in Fig.\ref{Fig7}, the change in parameters does not significantly impact the results, particularly for sharp features. However, when $N$ is fixed, a larger $\varphi$ value causes some non-feature points to be incorrectly identified as feature points (indicated by the green-dotted rectangles), whereas a smaller $\varphi$ value may lead to the omission of a few smooth features (highlighted by the red-dotted rectangles). Furthermore, when $\varphi$ is fixed, both larger and smaller values of $N$ lead to slight inaccuracies in feature identification. Despite this, most features are successfully captured, and the overall results remain satisfactory.

\subsection{Quantitative results}

In this section, we quantitatively evaluate the feature detection results by obtaining the outputs of 6000 models from the ABC dataset using our RWoDSN and the selected baselines. Table \ref{tab:table1} presents the results of the nine evaluation metrics mentioned earlier. As indicated by all metrics (except for precision), our method generates features that are much closer to the Ground Truth compared to the other methods. Although NerVE\cite{ref33}  achieves a higher precision, its recall is significantly lower than ours, indicating that many features are missed by NerVE\cite{ref33}, which is also confirmed by the qualitative results.

It is also evident that methods directly detecting feature points yield higher \textit{Jaccard}, \textit{IoU}, and \textit{MCC} scores compared to methods that fit edges. For example, these three metrics for NerVE\cite{ref33}, PBRG\cite{ref4} and SGLBP\cite{ref25} are higher than those for EC-Net\cite{ref9} and PIENet\cite{ref11}, as the former focus on detecting each individual point. Furthermore, the Hausdorff distance provides a global evaluation of how well the output features align with the Ground Truth, indicating that NerVE\cite{ref33} and EC-Net\cite{ref9} outperform traditional methods by effectively capturing the sharp features that define the overall shape of the models. However, EC-Net\cite{ref9} produces the lowest precision score, as it fits features rather than detecting them directly from the point cloud.  Regarding PIENet\cite{ref11}, it produces the lowest scores for \textit{Jaccard}, \textit{IoU}, \textit{MCC}, recall, \textit{MSE}, $d_{max}$ and $d_{min}$ scores, and we report different scores than those in \cite{ref11}, as PIENet\cite{ref11} focuses only on lines, circles, and B-spline curves, while neglecting "ellipse" and "other" types of feature curves. However, the 6000 models tested in this work contain diverse types of features, and the results from PIENet\cite{ref11} further verify the limited generalization ability of the curve-fitting strategy.

\begin{table}[H]
\caption{The quantitative results of the proposed methods and the baselines. The best scores are in \textbf{bold}, and the second-rank scores are underlined. \label{tab:table1}}
\centering
\begin{tabular}{llllllllll}
\hline
Methods & Hausdorff & \textit{Jaccard} & \textit{IoU} & \textit{MCC} & Precision & Recall & $MSE*100$ & $d_{max}$ & $d_{min}$\\
\hline
 
BE & 0.486 & 0.038 & 0.165 & 0.182 & 0.271 & 0.469 & 3.239 & 8.255 & 0.338\\
 
PBRG\cite{ref4} & 0.316 & 0.208 & 0.206 & 0.261 & 0.349 & 0.408 & 2.239 & 8.602 & 0.302 \\
 
SGLBP\cite{ref25} & 0.252 & 0.258 & 0.255 & 0.275 & 0.350 & 0.543 & 1.767 & 6.016 & 0.262 \\
 
EC-Net\cite{ref9} & 0.248 & 0.069 & 0.066 & -0.084 & 0.081 & 0.421 & 2.255 & 7.593 & \underline{0.181} \\

PIENet\cite{ref11} & 0.354 & 0.055 & 0.070 & 0.161 & 0.309 & 0.082 & 4.725 & 14.151 & 0.465 \\

NerVE\cite{ref33} & 0.252 & \underline{0.388} & \underline{0.388} & \underline{0.541} & \textbf{0.894} & 0.401 & 1.530 & 4.584 & 0.191\\

RWoDSN &\textbf{ 0.197} & \textbf{0.640} & \textbf{0.640} & \textbf{0.715} & \underline{0.784} & \underline{0.769} & \textbf{0.729} & \textbf{3.431} & \textbf{0.142} \\

RWoDSN(Noise\_1) & \underline{0.227} & 0.343 & 0.343 & 0.417 & 0.395 & \textbf{0.794} & \underline{0.990} & \underline{4.125} & 0.241 \\
 
RWoDSN(Noise\_3) & 0.228 & 0.329 & 0.329 & 0.371 & 0.380 & 0.775 & 1.098 & 4.290 & 0.227 \\
 
RWoDSN(Noise\_5) & 0.228 & 0.319 & 0.319 & 0.339 & 0.369 & 0.750 & 1.184 & 4.465 & 0.205\\ 
\hline
\end{tabular}
\end{table}

\subsection{Qualitative results}
\begin{figure}[H]
\centering
\includegraphics[width=1.0\linewidth]{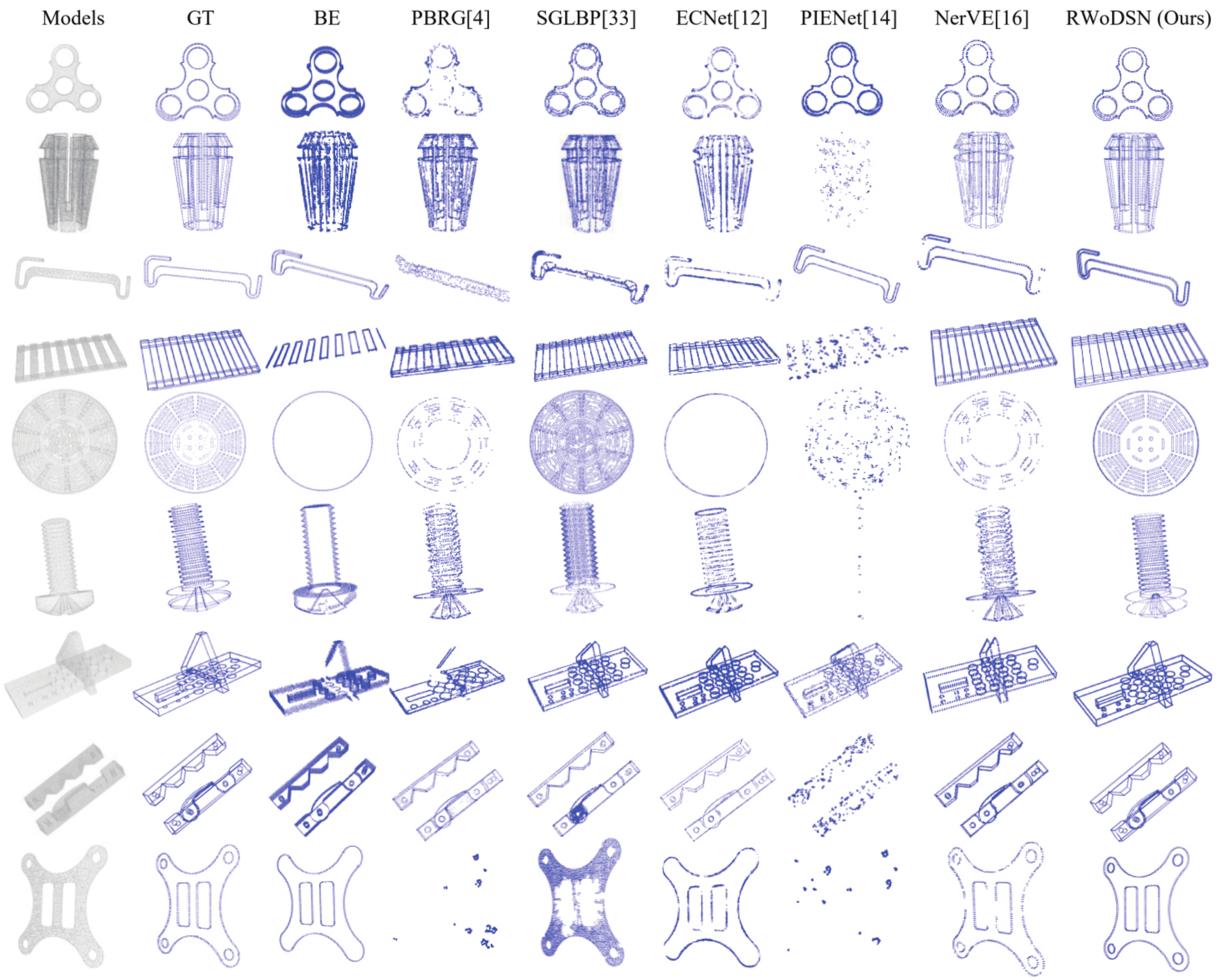}%
\caption{The results of the baseline methods compared with the proposed RWoDSN.}
\label{Fig8}
\end{figure}

The results of the proposed RWoDSN and the selected baselines on twelve typical models, which feature various types of features and exhibit non-uniform sampling patterns, are presented in Fig.\ref{Fig0} and Fig.\ref{Fig8}, respectively.  

From Fig.\ref{Fig0} and Fig.\ref{Fig8}, it is evident that the proposed RWoDSN significantly outperforms the other methods. In particular, BE is limited to extracting only the surface-like boundaries between two surfaces, leading to a higher number of redundant features. Both PBRG\cite{ref4} and SGLBP\cite{ref25} can capture most of the features; however, their results are not entirely satisfactory, as they tend to detect non-feature points that are close to the actual features or miss some smooth features. It is also noticeable that while EC-Net\cite{ref9} yields relatively satisfactory results, it fails to capture many features around the corners. PIENet\cite{ref11} produces the least satisfactory results, primarily because the input point clouds are down-sampled to 8096 points. While the down-sampled point clouds maintain the same number of points, they may not adequately represent the original surface, leading to reduced performance, especially on large-scale datasets. Furthermore, these twelve models contain diverse features beyond lines, circles, and B-spline curves. NerVE\cite{ref33} performs well both qualitatively and quantitatively but still struggles to handle fine-detailed features.

\begin{figure}[H]
\centering
\includegraphics[width=0.95\linewidth]{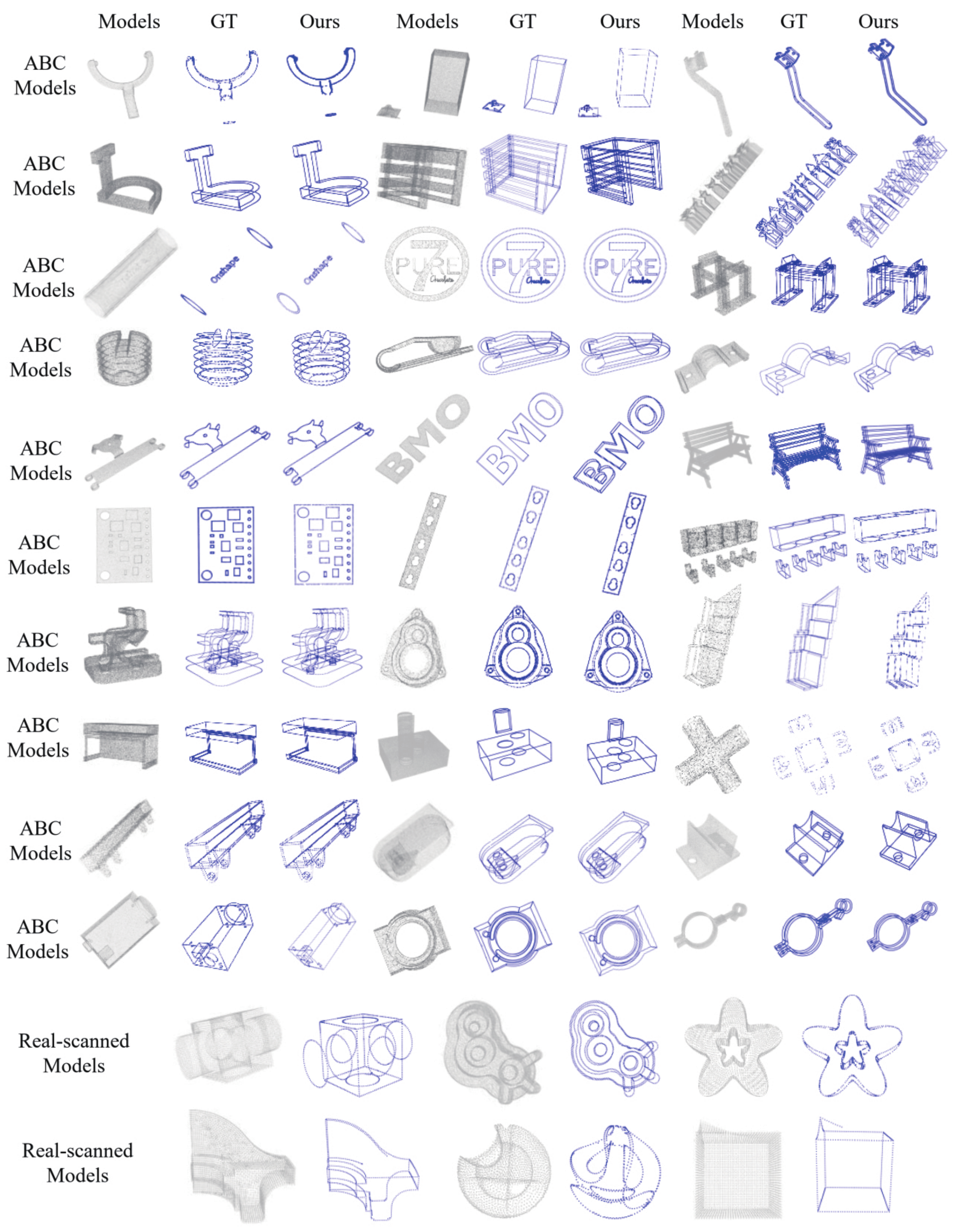}%
\caption{Qualitative results of our RWoDSN applied to several models from the ABC dataset and real-scanned models.}
\label{Fig9}
\end{figure}

We also present results on several models from the ABC dataset and some real-scanned models in Fig.\ref{Fig9}. The results shown in Fig.\ref{Fig9} demonstrate that RWoDSN effectively captures a wide range of features, from sharp to smooth and coarse to fine scales. This highlights its satisfactory performance across different models with varying types and scales of features. The proposed method detects features by analyzing the DSN, which encodes geometric, topological, and sampling information. Additionally, the constrained random walks on the DSN effectively mitigate the randomness inherent in the analysis, contributing to more accurate and reliable feature detection.

\subsection{Robustness}

\subsubsection{Robustness to noises} 

To test the robustness of our approach to noise, we evaluate the nine metrics at different noise levels with standard deviation of $\delta=0.01sd, 0.03sd, 0.05sd$. Here, $sd$ represents the sampling density $S_{density}$, computed using Equ.\eqref{S_density}, rather than the minimum value of sampling density, which provides a more rigorous stress test for the proposed approach.

The quantitative measures are listed in Table \ref{tab:table1}. As observed, our method still outperforms the other baselines in terms of Hausdorff distance, \textit{MSE}, and $d_{max}$ when tested on or trained on clean data. Moreover, our \textit{Jaccard}, \textit{IoU}, and \textit{MCC} scores remain very close to the best scores achieved on clean data. This is due to the fact that the vertices on the DSN represent the centroids of points within the same bin, rather than the original points on the point cloud. This approach helps to reduce the impact of noise disturbances to some extent. The results of RWoDSN on noisy models, shown in Fig.\ref{Fig10}, further demonstrate the robustness of our method to noise.

\subsubsection{Robustness to sampling density}

All the point clouds used in the experiments exhibit local imbalances in the sampling pattern, with a typical example shown in Fig.\ref{Fig0}. To further test the robustness of our method to varying sampling densities, we evaluated our approach on simplified models with simplification rates of $70\%$ and $50\%$, respectively. These simplified models represent the same surface but with different sampling densities. For a stress test, the simplification was performed using the RandomSample function implemented in 
PCL\cite{ref48}, rather than uniform sampling. 

\begin{figure}[H]
\centering
\includegraphics[width=1.0\linewidth]{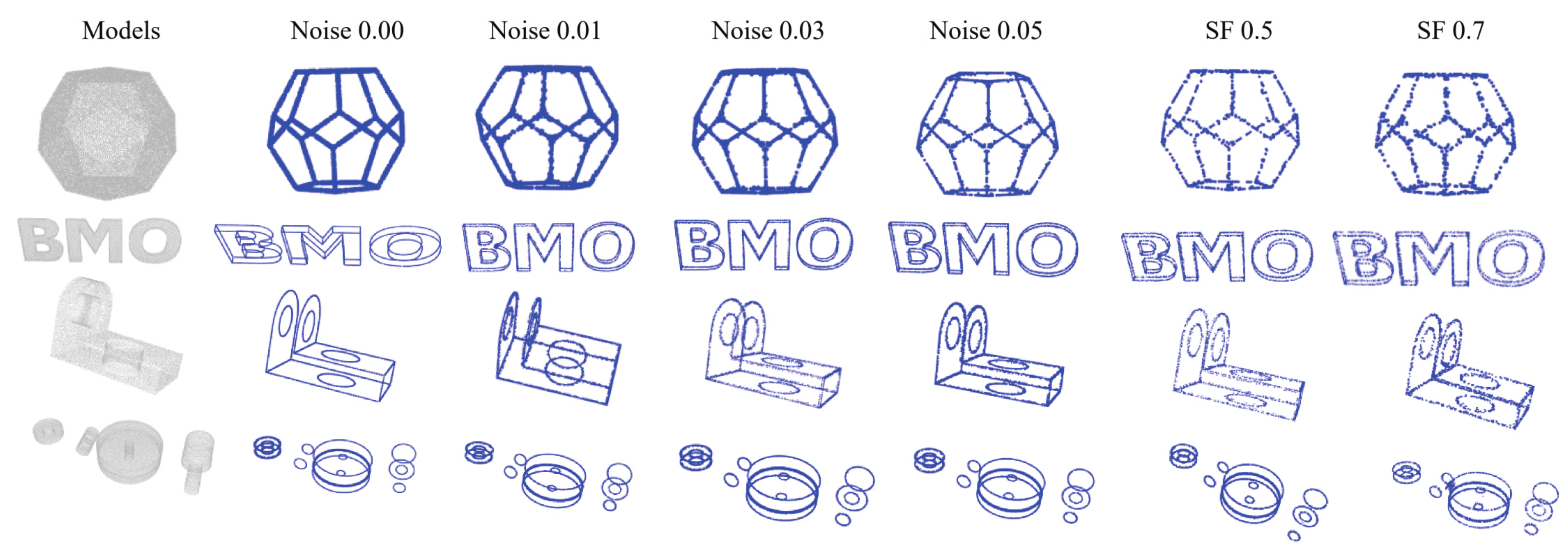}%
\caption{The results of our RWoDSN applied to noisy and simplified models.}
\label{Fig10}
\end{figure}

Since the nine evaluation metrics strongly depend on the number of points, which can result in misleading scores when applied to simplified models, we only present the qualitative results on the simplified models (shown in Fig.\ref{Fig10}). As demonstrated, our approach still successfully detects most of the features, even when processing simplified data. This robustness can be attributed to the fact that the vertices on the DSN represent the centroids of points within the same bin, thus mitigating the impact of variations in sampling densities.

\subsection{Feature-based surface segmentation}

\begin{figure}[H]
\centering
\includegraphics[width=0.7\linewidth]{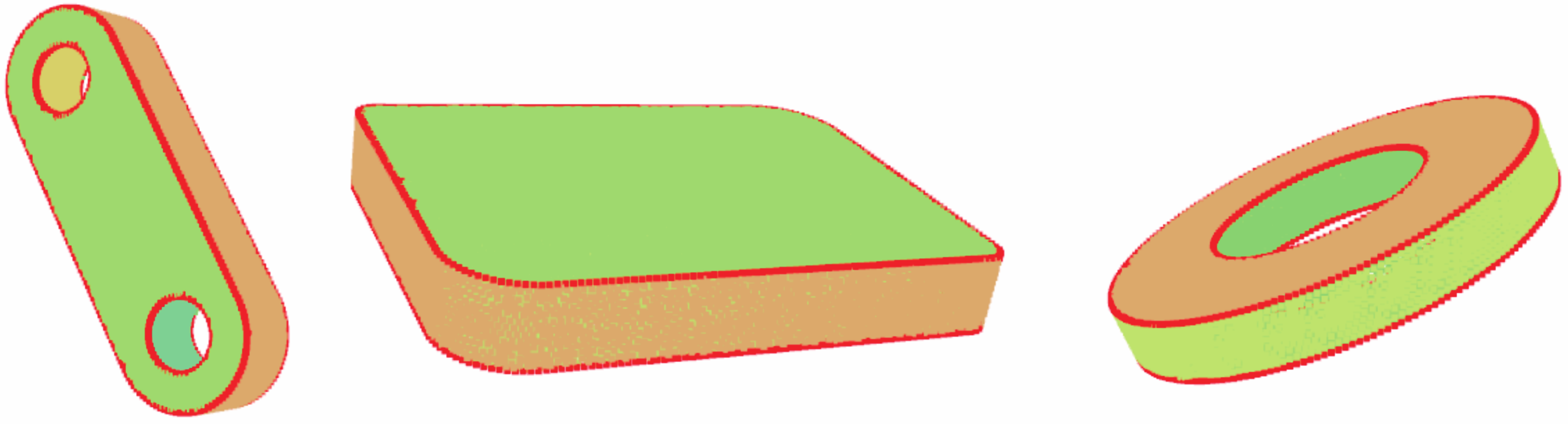}%
\caption{The results of surface segmentation.}
\label{SurfaceSeg}
\end{figure}

The proposed method accurately distinguishes feature points, enabling the extraction of different surfaces from the point cloud through a Breadth-First Search (BFS) starting from these feature points. Specifically, the BFS begins at a random, unvisited non-feature point (referred to as the seed point) and then explores the $k$NN ($k$ = 5) of the seed point. The BFS process terminates upon encountering a feature point and is repeated until no more seed points are found. Some results of surface segmentation are shown in Fig. \ref{SurfaceSeg}, further demonstrating the effectiveness and utility of the proposed method.

\subsection{Efficiency}

In this section, we compare the efficiency of the baseline methods with the proposed method. Since the input point clouds for PIENet are down-sampled to 8096 points, PIENet\cite{ref11} is not included in this comparison. All testing was performed on a system with the following specifications: GPU: NVIDIA RTX 2080; CPU: AMD Ryzen 7 3700X; and memory: 16 GB. For deep learning-based methods, we only report the testing time, excluding the training time.

We tested the efficiency of the methods using five different models, with scales ranging from 5,192 to 1,357,762 points. The detailed timing information is provided in Table \ref{tab:table2}. As shown in the table, the proposed method takes more time than BE. However, compared with PBRG\cite{ref4}, the proposed method is markedly more efficient, as PBRG\cite{ref4} involves multiple visits to some features, resulting in higher computational costs. Additionally, while the efficiency of SGLBP\cite{ref25} and RWoDSN is similar for small-scale models, SGLBP\cite{ref25} becomes significantly slower when dealing with models containing millions of points. 

Deep-learning-based methods generally exhibit better efficiency than traditional methods. For instance, NerVE\cite{ref33} and EC-Net\cite{ref9} outperform all traditional methods in terms of efficiency. However, these methods require more computational resources and training with supervised labels. Although the efficiency of RWoDSN is not optimal due to the multiple iterations of constrained random walks, its feature detection results significantly surpass those of other methods, as demonstrated in the previous sections. This indicates that the proposed method is particularly suitable for applications where accuracy is prioritized over efficiency.

\begin{table}[H]
\caption{The timings of the baseline methods and the proposed RWoDSN. “-” indicates that no results were obtained.} 
\label{tab:table2}
\centering
\begin{tabular}{clllllll}
\hline
Point & \multirow{2}{*}{$\#$Points} & \multicolumn{6}{c}{Runing Times(s)} \\

\cline{3-8}

Cloud  &  & RWoDSN & BE & PBRG\cite{ref4} & SGLBP\cite{ref25} & EC-Net\cite{ref9} & NerVE\cite{ref33}\\ 
\hline

Point cloud \#1 & 5192 & 1.274 & 1.010 & 9.063 & 1.003 & 1.527 & 0.118\\

Point cloud \#2 & 19398 & 13.600 & 2.130 & 176.792 & 5.891 & 5.297 & 0.113\\

Point cloud \#3 & 104412 & 305.189 & 61.791 & 17745.690 & 659.166 & 30.477 & 0.168\\

Point cloud \#4 & 564776 & 10249.500 & 16815.720 & -– & 8789.617 & 170.304 & 0.488\\

Point cloud \#5 & 1357762 & 23005.165 & 242.561 & -– & 53568.204 & 305.431 & 0.134\\
\hline
\end{tabular}
\end{table}

\subsection{Limitations}

The proposed RWoDSN may encounter difficulties when handling extremely sparse models, as the resampling process may fail to obtain centroids in the corresponding bins. Additionally, the main drawback of the proposed RWoDSN is its efficiency, as the random walk process is performed multiple times for each point, resulting in higher computational costs.

\section{Conclusion}
In this work, we have proposed a novel feature detection method called RWoDSN. Our approach first constructs the DSN descriptor for the neighborhood of a point and then performs the random walk process on the DSN. As a result, a graph-based DSN is obtained to identify the location of a point, distinguishing feature points from non-feature points. Experiments demonstrated that our method achieves a recall of 0.769, which is 22\% higher than the state-of-the-art, with a precision of 0.784, and produces the best scores across eight metrics. Furthermore, qualitative results indicate that the proposed RWoDSN outperforms previous methods in terms of accuracy. Additionally, the experiments show that the effectiveness of our method is not significantly sensitive to variations in each parameter and that it is robust to noise and imbalanced sampling densities.

Moreover, the experimental results reveal that fitting feature strategies generally perform better in capturing the global shape of the models, while point processing strategies achieve higher scores in point identification metrics. In future work, we aim to combine the DSN with deep networks, extending RWoDSN into deep learning strategies to enhance both efficiency and the ability to learn from a wider range of features.

\bibliographystyle{model3-num-names} 
\bibliography{ref}

\end{document}